\documentclass[a4paper,conference]{IEEEtran}

\usepackage{url}
\usepackage{amssymb}
\usepackage{lineno}
\usepackage{tikz}
\usepackage{pgfplots}
\usepackage[utf8]{inputenc}
\usepackage{graphicx}
\usepackage{float}
\usepackage{listings}
\usepackage{verbatim}

\usepackage[shortcuts,acronym]{glossaries}
\usepackage{color}
\usepackage{comment}
\usepackage{hyperref}
\usepackage[utf8]{inputenc}

\usepackage[acronym]{glossaries}
\makeglossaries
\newcommand{\todo}[1]{\textcolor{red}{TODO: #1}}

\definecolor{mycolor1}{rgb}{0.00000,0.44700,0.74100}
\definecolor{mycolor2}{rgb}{0.85000,0.32500,0.09800}
\definecolor{mycolor3}{rgb}{0.92900,0.69400,0.12500}

\newacronym{CT}{CT}{Computed Tomography}
\newacronym{FH}{FH}{Feature Histograms}
\newacronym{RNA}{RNA}{Ribonucleic Acid}
\newacronym{COVs}{COVs}{Coronaviruses}
\newacronym{SARS}{SARS}{Severe Acute Respiratory Syndrome}
\newacronym{ARDS}{ARDS}{Acute Respiratory Distress Syndrome}
\newacronym{MERS}{MERS}{Middle East Respiratory Syndrome}
\newacronym{WHO}{WHO}{World Health Organization}
\newacronym{AI}{AI}{Artificial Intelligence}
\newacronym{RT-PCR}{RT-PCR}{ Real-time Reverse Transcription Polymerase Chain Reaction}
\newacronym{RCNN}{RCNN}{Recurrent Convolutional Neural Network}
\newacronym{BCL}{BCL}{Bilateral Convolution Layer}
\newacronym{USM}{USM}{Unsharp Masking}
\newacronym{PC}{PC}{PlantCLEF}
\newacronym{NLM}{NLM}{Non Local Means}
\newacronym{DNLM}{DNLM}{Deceived Non Local Means}
\newacronym{SVM}{SVM}{Support Vector Machine}
\newacronym{LoG}{LoG}{Laplacian of Gaussian}
\newacronym{DoG}{DoG}{Difference of Gaussians}
\newacronym{SP}{SP}{Salt and Pepper}
\newacronym{RPN}{RPN}{Region Proposal Network}
\newacronym{ReLU}{ReLU}{Rectified Linear Unit}
\newacronym{FCN}{FCN}{Fully Convolutional Network}
\newacronym{ROI}{ROI}{Region of Interest}
\newacronym{DDT}{DDT}{Deep Distance Transformer}
\newacronym{Top-WS}{Top-WS}{Watersheds instance segmentation as top-model}
\newacronym{Top-RPN-WS}{Top-RPN-WS}{Region proposal network and watersheds top-model}
\newacronym{Top-Unet3}{Top-Unet3}{U-net top-model}
\newacronym{Top-Unet-ML}{Top-Unet-ML}{U-net with modified loss function top-model}
\newacronym{U-net3}{U-net3}{U-net for three classes}
\newacronym{Unet-ML}{Unet-ML}{U-net with the proposed modified loss function}
\newacronym{DTGT}{DTGT}{Distance Transform Ground-truth}
\newacronym{MCD}{MCD}{Monte Carlo Dropout}
\newacronym{BTGT}{BTGT}{Border Transform Ground-truth}
\newacronym{WDMC}{WDMC}{Weighted Dice Multiclass Coefficient}
\newacronym{BDE}{BDE}{Boundary Displacement Error}
\newacronym{MAE}{MAE}{Mean Absolute Error}
\newacronym{MSE}{MSE}{Mean Squared Error}
\newacronym{PSP}{PSP}{photostimulable phosphor}
\newacronym{CMOS}{CMOS}{Complementary Metal-oxide-semiconductor}
\newacronym{CMOS-APS}{CMOS-APS}{Complementary Metal Oxide Semiconductor Active Pixel Sensor}
\newacronym{Grad-CAM}{Grad-CAM}{Gradient Class Activation Maps}
\newacronym{PACS}{PACS}{picture archiving and communication system}
\newacronym{TIFF}{TIFF}{tagged image file format}
\newacronym{MRI}{MRI}{Magnetic Resonance Imaging}
\newacronym{CNN}{CNN}{Convolutional Neural Network}
\newacronym{CAD}{CAD}{Computer Aided Diagnosis}
\newacronym{OOD}{OOD}{Out of Distribution}
\newacronym{DNN}{DNN}{Deep Neural Networks}
\newacronym{IOD}{IOD}{In-Distribution}
\newacronym{GAN}{GAN}{Generative Adversarial Network}
\newacronym{DeDiM}{DeDiM}{Deep Dataset Dissimilarity Measure}
\newacronym{PBC}{PBC}{Pseudo-label based Balance Correction}

\newacronym{SSDL}{SSDL}{Semi-supervised Deep Learning}

\newacronym{ROC}{ROC}{Receiver Operator Characteristic}

\newacronym{UDA}{UDA}{Unsupervised Domain Adaptation}
\newacronym{TB-1}{TB-1}{Test-bed 1}
\newacronym{TB-1.1}{TB-1.1}{Test-bed 1.1}
\newacronym{TB-1.2}{TB-1.2}{Test-bed 1.2}
\newacronym{TB-2}{TB-2}{Test-bed 2}

\begin{document}

\title{Dealing with Distribution Mismatch in Semi-supervised Deep Learning  for Covid-19 Detection Using Chest X-ray Images: A Novel Approach Using Feature Densities}

\author{\IEEEauthorblockN{Saul Calderon-Ramirez \IEEEauthorrefmark{1}\IEEEauthorrefmark{2}, Shengxiang Yang \IEEEauthorrefmark{1}, \IEEEauthorblockN{David Elizondo \IEEEauthorrefmark{1}, Armaghan Moemeni \IEEEauthorrefmark{6}}
 \vspace{0.25cm}}

\IEEEauthorblockA{\IEEEauthorrefmark{1} Centre for Computational Intelligence (CCI), De Montfort University, United Kingdom}

\IEEEauthorblockA{\IEEEauthorrefmark{6}School of Computer Science, University of Nottingham, United Kingdom}

\IEEEauthorrefmark{1}sacalderon@itcr.ac.cr,

\IEEEauthorrefmark{1}syang@dmu.ac.uk,
\IEEEauthorrefmark{2}armaghan.moemeni@nottingham.ac.uk, \\

\IEEEauthorrefmark{1}elizondo@dmu.ac.uk,

}

\maketitle

\begin{abstract}
In the context of the global coronavirus pandemic, different deep learning solutions for infected subject detection using chest X-ray images have been proposed. However,  deep learning models usually need large labelled datasets to be effective.  Semi-supervised deep learning is an attractive alternative, where unlabelled data is leveraged to improve the overall model's accuracy. However, in real-world usage settings, an unlabelled dataset might present a different distribution than the labelled dataset (i.e. the labelled dataset was sampled from a \textit{target} clinic  and the unlabelled dataset from a \textit{source} clinic). This results in a distribution mismatch between the unlabelled and labelled datasets. In this work, we assess the impact of the distribution mismatch between the labelled and the unlabelled datasets, for a semi-supervised model trained with chest X-ray images, for COVID-19 detection.  Under strong distribution mismatch conditions, we found an accuracy hit of almost 30\%, suggesting that the unlabelled dataset distribution has a strong influence in the  behaviour of the model. Therefore, we propose a straightforward approach to diminish the impact of such distribution mismatch. Our proposed method uses a density approximation of the feature space. It is  built upon the target dataset  to filter out the observations in the source unlabelled dataset that might harm the accuracy of the semi-supervised model.  It assumes that a small labelled source dataset is available together with a larger source unlabelled dataset. Our proposed method does not require any model training, it is simple and  computationally cheap. We compare our proposed method against two popular state of the art \textit{out-of-distribution}  data detectors, which are also cheap and simple to implement. In our tests, our method yielded accuracy gains of up to 32\%, when compared to the previous state of the art methods. The good results yielded by our method leads us to argue in favour for a more data-centric approach to improve model's accuracy. Furthermore, the developed method can be used to measure data effectiveness for semi-supervised deep learning model training.

\end{abstract}

\section{Introduction}
The  COVID-19 disease is caused  by the novel SARS-CoV2 coronavirus, discovered in 2019 \cite{sun2020covid}.
The COVID-19 pandemic has caused thousands of human losses around the world, where even the most developed health systems have not been able to cope with the infection peaks \cite{sun2020covid}. Health practitioners are struggling with the detection and tracking of infected subjects, as the number of patients in need for medical assistance increases. 

Therefore, accurately detecting patients infected with the SARS-CoV2 virus is a critical task to control the pandemic. Nevertheless, SARS-CoV2  detection methods like the \gls{RT-PCR} test can be expensive and time consuming. As an alternative and/or complementary method, the usage of medical imaging based approaches can be less expensive and also accurate \cite{chen2020epidemiological,chung2020ct}. Moreover, X-ray based imaging diagnosis can be considered cheaper. The usage of X-ray machines is more widespread when compared to other imaging technologies like computer tomography. This is  specially the case in less industrialised countries \cite{Arora2014}. However, a limitation of X-ray based diagnosing of COVID-19 is the need of highly trained clinical practitioners like radiologists, which in less industrialised countries are scarce \cite{Arora2014}. 

The implementation of \gls{CAD} systems for COVID-19 diagnosis can be a solution to mitigate the specialized staff shortage. Deep learning based \gls{CAD} systems have been extensively explored for different medical imaging applications \cite{bermudezquality,calvo2019assessing,alfaro2019brief,calderon2021real,zamora2021enforcing,calderon2021improving}. More specifically, several deep learning architectures for COVID-19 detection have been proposed recently in the literature \cite{ismael2021deep,jain2021deep,basu2020deep}. These systems have been developed using publicly available X-ray images datasets, with COVID-19 positive  \cite{cohen2020covid} and negative cases \cite{calderon2021dealing}.  

Nevertheless, a short-coming of implementing a deep learning architecture for real-world usage  is the need of a large labelled dataset from the specific target clinic or hospital where the system is intended to be used. Labeling images in the medical domain is time-consuming and requires expensive human effort from highly trained clinical practitioners, which makes building an extensive labelled dataset costly. Previous work on COVID-19 detection with deep learning has relied on large and heteregenous datasets, where around 100-400 COVID-19 positive cases sampled from the dataset \cite{cohen2020covid}, and larger datasets of COVID-19 negative cases sampled from different sources \cite{kermany2018identifying,irvin2019chexpert,demner2016preparing}. Such testing conditions can be considered far from a real-world scenario, where usually in the target clinic/hospital a limited set of labelled observations is available. Using external datasets for training might harm the overall performance  of the model. This is mainly due to the differences between patient features and imaging protocols. This affects the final data distribution between the test and training data  \cite{tan2019semi}. 

Another short-coming of the aforementioned  previous work, is the bias of the population between the positive and negative COVID-19 samples. For example, as reported in \cite{roberts2021common}, negative COVID-19 observations in \cite{kermany2018identifying} were sampled from pediatric chinese patients, while positive COVID-19 cases in \cite{cohen2020covid} correspond to adult patients from different countries. This dataset combination has been extensively used for training \gls{CNN} based models to detect COVID-19, and leads to deceptive bias in both the test and training model data \cite{roberts2021common}. 

To deal with the limited labelled datasets, different approaches have been implemented in literature \cite{cheplygina2019not}. In the context of COVID-19 detection, namely data augmentation and transfer learning \cite{maghdid2021diagnosing,elgendi2021effectiveness} have been used. In transfer learning, a source labeled dataset $D^{s}_l$ is used to pre-train a model, and then fine-tune it in the target dataset $D^{t}_l$. However, as discussed in \cite{zhou2021soda}, fine-tuning might not be enough to improve the model's accuracy.  The distribution mismatch between $D^{s}_l$ and $D^{t}_l$ due to different patient populations and imaging acquisition protocols, is frequently a reason for poor transfer learning performance.

Another approach to deal with scarce labelled data is the usage of \gls{SSDL}. \gls{SSDL} leverages cheaper and more widely available unlabelled data. Semi-supervised learning for COVID-19 detection have been explored in \cite{calderon2021dealing,calderon2020correcting} with positive results, where very small labelled datasets have been used. The authors combined \gls{SSDL} with common data augmentation and transfer learning approaches.  However, to implement deep learning based solutions for extensive real-world usage, testing different model attributes like robustness and predictive uncertainty is crucial for its safe usage.  A deep review on the importance of measuring different model attributes like robustness in medical applications of  \gls{AI} can be found in \cite{oala2020ml4h}. In a real-world scenario, the use of unlabelled data sampled from different sources (hospitals or clinics) can be considered. However, the usage of unlabelled datasets with different distributions from the labelled test and training target data might harm the accuracy of the model. This leads to the need of analyzing model robustness to different data distributions in the unlabelled dataset.  Therefore, in this work, we study the impact of different unlabelled data sources in a \gls{SSDL} model. Specifically, the MixMatch algorithm, which previously yielded interesting accuracy gains with very small labelled datasets for COVID-19 detection using X-ray images \cite{calderon2020correcting,calderon2021dealing} is used. Moreover, we propose a simple approach to select and build an unlabelled dataset. This aims to improve the overall \gls{SSDL} model accuracy.
Therefore, in this work, we evaluate a setting where the following datasets are available:

\begin{enumerate}
    \item A labelled dataset in the target clinic/hospital $D^{l}_t$ is available. The number of labelled observations $n^{l}_t$ is very small.  The target dataset is sampled from the clinic/hospital where the model is intended to be deployed. 
    \item A larger unlabelled dataset in a different source clinic/hospital $D^{u}_s$ is available, with  $n^{u}_s>n^{l}_t$.
\end{enumerate}

Different  deep learning applications in medical imaging face distribution mismatch situations between the different datasets used. This might be the case for \gls{SSDL}, when using different unlabelled data sources. We argue that quantifying distribution mismatch with respect to the model behaviour is important for medical imaging applications, as different unlabelled data sources might be considered.  Moreover, simple dataset transformation procedures to improve model robustness to data distribution mismatch between the labelled and unlabelled datasets, is also important. This helps to narrow the gap between machine learning research and its real-world usage.

\section{State of the art}\label{sec:stateoftheart}

\subsection{Semi-supervised Deep Learning} 
\gls{SSDL} aims to deal with small labelled datasets, by leveraging unlabelled data.  Supervised deep learning networks often require large labelled datasets. This is partially addressed with the usage of data augmentation and transfer learning \cite{van2001art}. However, the usage of cheaper and more widely available unlabelled data, can further lower the need for labelled data.  With a formal notation, in \gls{SSDL}  both labelled and unlabelled datasets are used. Each labelled observation $X_{l}=\left\{ \boldsymbol{x}_{1},\ldots,\boldsymbol{x}_{n_{l}}\right\}$ is mapped to a label in the set $Y_{l}=\left\{ y_{1},\ldots,y_{n_{l}}\right\} $. The unlabelled dataset corresponds to a set of observations  $X_{u}=\left\{ \boldsymbol{x}_{1},\ldots,\boldsymbol{x}_{n_{u}}\right\} $, with $S_u = X_u$.

\gls{SSDL} architectures can be classified as: Pre-training  \cite{doersch2015unsupervised}, pseudo-labelled  \cite{dong2018tri} and regularization based. Within regularization based approaches, consistency loss term and graph based regularization and generative based  \cite{cheplygina2019not} regularization techniques can be distinguished. A detailed survey regarding \gls{SSDL} can be found in \cite{van2020survey,kim2021recent}. 

Concerning regularization based \gls{SSDL}, a regularization term leveraging unlabelled data is implemented in the loss function $S_{u}$: 
\begin{equation}
\mathcal{L}\left(S\right)=\sum_{\left(\boldsymbol{x}_{i},\boldsymbol{y}_{i}\right)\in S_{l}}\mathcal{L}_{l}\left(\boldsymbol{w},\boldsymbol{x}_{i},\boldsymbol{y}_{i}\right)+\gamma\sum_{\overrightarrow{x}_{j}\in X_{u}}\mathcal{L}_{u}\left(\boldsymbol{w},\boldsymbol{x}_{j}\right),
\label{eq:reguLearning}
\end{equation}
with $\boldsymbol{w}$  the model's  weights array, $\mathcal{L}_{l}$ and $\mathcal{L}_{u}$ the labelled and unlabelled loss terms respectively. The  coefficient $\gamma$ weighs the influence of unsupervised regularization. As previously mentioned, a number of regularization based variations can be found in the literature. The main ones include: consistency loss based  \cite{tarvainen2017mean,tan2019semi}, graph based \cite{weston2012deep,luo2018smooth} and generative augmentation based \cite{springenberg2015unsupervised,salimans2016improved}. Consistency based methods make the assumption of clustered-data/low-density separation. Such assumption refers to how the observations corresponding to a class, are clustered together. This makes the decision manifold lie in very sparse regions \cite{van2020survey}. A violation to this assumption might degrade the performance of the semi-supervised method \cite{van2020survey}.

In pseudo-label training, pseudo-labels are estimated
for unlabelled data. These are used for later model refinement. A straightforward pseudo-label based approach is based in co-training two models \cite{balcan200621}. The model is pre-trained with the limited size labelled dataset. Later, the pseudo-labels are estimated for the unlabelled data using two models trained with different views (features) of the data. A voting scheme is implemented for estimating the pseudo-labels. 

Mix Match \cite{berthelot2019mixmatch} combines both pseudo-label and  consistency based \gls{SSDL}, along  with heavy data augmentation using the MixUp algorithm \cite{zhang2017mixup}. According to  \cite{berthelot2019mixmatch}, MixMatch out-performs, accuracy wise, previous \gls{SSDL} approaches. Given the recently state of the art performance demonstrated by Mix Match and also the good results yielded in \cite{calderon2021dealing,calderon2020correcting} for medical imaging applications, we chose it for the developed solution in this work. A detailed description of MixMatch can be found in Section \ref{sec:Proposed_method}.

\subsection{\gls{SSDL} robustness to distribution mismatch}

The distribution mismatch between $S_u$ and $S_l$ is  also referred to as the identically and independently distributed (IID) assumption violation. It might have different degrees and causes, which are enlisted as follows \cite{kairouz2019advances}:
\begin{itemize}
    
    \item Prior probability shift: The distribution of the labels in $S_l$ can be different when compared to $S_u$. In a \gls{CAD} system this can be exemplified when the labels of the medical images have different distributions between the two datasets $S_l$ and $S_u$. A specific case would be the label imbalance of the labeled dataset $S_l$ as discussed in \cite{calderon2020correcting}.
    \item Covariate shift: A different distribution of the features in the input observations might be sampled, leading to a distribution mismatch. In a medical imaging application, this can be related to the difference in the  frequencies of the observed features between $S_l$ and $S_u$.
    \item Concept drift: It refers to the different features observed in a sample, with the same label. In the application at hand in this work, this might happen when different patients with different variations of the COVID-19 disease are sampled to build $S_u$ with the same pathologies (classes) in $S_l$.
    \item Concept shift: It is associated to a shift in the labels, with the same features. In the aforementioned example, it would refer to labelling a medical image with similar features with a different pathology (a bias caused by the image labelers). 
    
\end{itemize}

In our tested setting, different data sources were used only to gather unlabelled data $S_u$. We recreate two of the aforementioned distribution mismatch causes: covariate and prior probability shift. The unlabelled datasets created and tested belong to normal (no pathology) chest X-ray images (COVID-19$^{-}$), from patients of different nationalities. As the labelled dataset $S_l$ includes both classes (COVID-19$^{+}$ and COVID-19$^{-}$), a label distribution mismatch also occurs. The tested setting in this work simulates the case where different unlabelled data sources might be available (for instance from different hospitals), at the beginning of a pandemic. Furthermore, a small labelled dataset might be available in the target hospital/clinic. 

The usage of different unlabelled datasets might potentially cause a violation of the aforementioned clustered-data/low-density separation assumption. Using unlabelled datasets with different  distributions when compared to the labelled dataset, might create wrong sparse regions and/or less clustered groups of observations belonging to the same class. Therefore, in this work we explore data-oriented approaches to deal with potential violations of the clustered-data/low-density separation assumption. Unlabelled data can be considered significantly cheaper than labelled data. Thus, discarding potentially harmful  observations with the aim to decrease the odds of violating the clustered-data/low-density separation assumption  is  viable and worthy to explore.

In \cite{oliver2018realistic}, an extensive evaluation of different distribution mismatch settings and its impact in \gls{SSDL} is developed. Authors concluded that distribution mismatch in \gls{SSDL} is an important challenge to be addressed. Recently, different approaches for improving \gls{SSDL} robustness to the distribution mismatch between $S_u$ and $S_l$ have been proposed. In \cite{nair2019realmix}, an \gls{OOD} masking method is proposed. It consists on weighting the observations likely to be \gls{OOD} during semi-supervised training. The output of a softmax activation function after the raw model output, was used as \gls{OOD} masking coefficient. This works as an observation-wise weighting during semi-supervised model training. The authors compared their proposed method with state of the art general-purpose \gls{SSDL} approaches like MixMatch \cite{berthelot2019mixmatch}. The test bed consisted in different unlabelled datasets with a varying degree of distribution mismatch. The contamination source consists of images with different labels and features (completely \gls{OOD}). Their method proved to improve model robustness against \gls{OOD} data contamination in $S_u$, using general purpose datasets such as CIFAR-10 and SVHN. However, other types of distribution mismatch corruption such as concept drift or covariate shift were not tested. 

Another approach to deal with distribution mismatch under \gls{OOD} contamination (different labels and features), can be found in \cite{chen2020semi}. The proposed method also implements a weighting coefficient, calculated as the softmax output of a models ensemble. In a similar trend, the work in \cite{guo2020safe} propose a weighted approach to deal with \gls{OOD} observations (with different label, different features). However, instead of using the softmax output, the observation-wise weight is estimated through an optimization step. Similar to \cite{nair2019realmix}, only general purpose datasets (CIFAR-10 and MNIST)  were used, with no other variations of distribution mismatch settings. Another resembling approach and testing bed to \cite{guo2020safe}, can be found in \cite{zhao2020robust}, where an optimization based approach to weight each observation is implemented, with a test-bed  focused in \gls{OOD} contaminated unlabelled datasets.  


In this work, we analyze the effect of distribution mismatch in \gls{SSDL} within a real-world application: COVID-19 detection using chest X-ray images. Unlike previous work on \gls{SSDL} under distribution mismatch, we test a real-world setting in the medical domain, and explore its implications within such context. As previously mentioned, we analyze the impact of a distribution mismatch caused by covariate and prior probability shift. Different unlabelled dataset sources within the same domain and features are used. We aim to evaluate different approaches to weigh how harmful an unlabelled observation could be for \gls{SSDL} training. We test different \gls{OOD} detection approaches in this work. After calculating a \textit{harm}
 coefficient for each unlabelled observation, different steps can be implemented to use such unlabelled dataset. For example,  filtering the observations with high \textit{harm} coefficients, select an unlabelled dataset upon its estimated benefit for \gls{SSDL}, or weigh the unlabelled observation  during \gls{SSDL} training.

Moreover, we focus on a data-oriented approach to identify and/or build a good unlabelled dataset for \gls{SSDL}. We propose a simple and very inexpensive method to evaluate the distribution mismatch between an unlabelled and labelled datasets, $S_u$ and $S_l$ respectively. Such method can be thought  as an \gls{OOD} scoring approach (\textit{harm} coefficient), which leads us to compare our method to recent \gls{OOD} detectors used in the context of \gls{OOD} data filtering to improve the accuracy of an \gls{SSDL} model. 

\subsection{OOD data detection}

\gls{OOD} data detection refers to the general problem of detecting observations that are very unlikely given a specific data distribution (usually the training dataset distribution) \cite{hendrycks2016baseline}.  The problem of \gls{OOD} data detection can be thought  as a generalization of the outlier detection problem, as it considers individual and collective outliers \cite{singh2012outlier}. Specific scenarios of \gls{OOD} data detection can be found in the literature. These include novel data and anomaly detection \cite{perera2019deep}, with several applications like rare event detection \cite{hamaguchi2019rare,amodei2016concrete}. In classical pattern recognition literature different approaches to anomaly and \gls{OOD} data detection are grounded in concepts such as density estimation \cite{markou2003novelty}, kernel representations \cite{10.1023/B:MACH.0000008084.60811.49}, prototyping \cite{markou2003novelty} and robust moment estimation \cite{doi:10.1080/01621459.1984.10477105}.

Recent success of deep learning based approaches for image analysis  \cite{wason2018deep} have motivated the development of \gls{OOD} detection techniques for deep neural networks. \gls{OOD} detection methods with deep learning architectures can be categorized in methods based upon the \gls{DNN}'s output, its input, or its learned feature space.

\gls{DNN}'s output based methods include the softmax based \gls{OOD} detector proposed in \cite{DBLP:journals/corr/HendrycksG16c}. In such work, \gls{OOD} detection is framed as a confidence estimation using the model's raw output layer values and passing it through a softmax function. Its maximum softmax value is used as confidence. Authors claim that the highest softmax  value of \gls{OOD} observations meaningfully differ from in distribution observations.

However, as reported in \cite{liang2018enhancing}, non calibrated models can be overconfident with \gls{OOD} data. Therefore, in \cite{liang2018enhancing} a calibration methodology is introduced, implementing a temperature coefficient. \gls{OOD} data detection  in neural networks is implemented in \cite{liang2018enhancing} using input perturbations meant to maximize the softmax based separability. For this  end, a gradient descent optimization is used, resulting in a preprocessed image.  A \emph{temperature} coefficient   in the calculation of the softmax output is added and is estimated to make the true positive rate of 95\% for in-distribution data detection, using the previously pre-processed images.  

Another approach for \gls{OOD} detection based on the model's output is the usage of  \gls{MCD} based uncertainty estimations.\gls{MCD} is a popular method for implementing predictive uncertainty estimation \cite{loquercio2020general,kendall2017uncertainties}. It consists in analyzing the  distribution  of $N$ predictions using the same input and adding noise to the model (drop-out in the context of \gls{DNN}s). This idea has been ported to the \gls{OOD} detection problem, where observations with high uncertainty are scored with high \gls{OOD} likelihood \cite{jin2019augmenting,sedlmeier2020uncertainty}. 

Regarding  feature space (a latent space approximation in \gls{DNN}s) based methods for \gls{OOD} detection different approaches can be found in the literature.  For example, in \cite{lee2018simple}, the authors implemented  the Mahalanobis distance in latent space of the dataset to the input observation, assuming a Gaussian distribution of the data. Both the mean and covariance  are estimated for the in distribution dataset. For a new observation $\boldsymbol{x}$, the \gls{OOD} score is estimated as the Mahalanobis distance for such given distribution.  The authors also implemented the calibration approach used in \cite{liang2018enhancing}. A superior performance of their proposed method in generic \gls{OOD} detection benchmarks is reported, when compared to the methods in \cite{liang2018enhancing,DBLP:journals/corr/HendrycksG16c}. However, no statistical significance tests of the results were performed. 

Another feature space based approach can be found in \cite{vanuncertainty}, known  as deterministic uncertainty quantification. Such approach is also intended for uncertainty estimation, but also is tested as an \gls{OOD} detection technique. It makes use of a centroid calculation of each category in the feature space, to later quantify the distance of a new observation to each centroid. Uncertainty quantification is estimated based in the kernel based distance to the category centroids. The approach is compared against an ensemble of deep neural networks (an output based approach for \gls{OOD} detection). This is done in a simple \gls{OOD} detection benchmark, where the CIFAR-10 is used as an in-distribution dataset and the SVHN as a \gls{OOD} dataset.  The authors reported the area under the \gls{ROC} curve  of their approach against other \gls{OOD} methods. Their approach showed the highest area under the \gls{ROC} curve index. However, no statistical analysis of the results were done. 

In \cite{calderonramirez2020mixmood} the authors developed an extensive testing of the influence of distribution mismatch between unlabelled and labelled datasets. Moreover, they also developed an approach to estimate the accuracy hit of such distribution mismatch for a state of the art \gls{SSDL} method. The proposed method estimates the distribution mismatch in the feature space between $S_l$ and $S_u$, using what the authors referred as a \gls{DeDiM}. Euclidean and Manhattan based \gls{DeDiM}s were tested and compared against density based \gls{DeDiM}s. All of them were applied within the feature space, built with an image net pre-trained network. The authors found a significant advantage of the density based distances.  In \cite{zisselman2020deep}, the authors proposed an \gls{OOD} detector using the feature space as well. The approach fits different parametric distributions in the feature space of the data. The decision to discriminate between \gls{OOD} and \gls{IOD} data is done based on the estimation of the approximated parametric model. Unfortunately, no comparison with other popular \gls{OOD} methods was presented.

\subsubsection{Unsupervised Domain Adaptation}
When using an unlabelled dataset $S_u$ with a very different distribution to $S_l$, a solution would be to \textit{correct} or  \textit{align} the feature extractor trained with labelled or unlabelled data from the source of the unlabelled dataset $S_u$, to the distribution of the labelled dataset  $S_l$ (target dataset, usually smaller). This is known as \gls{UDA}. For instance in \cite{zhou2021soda}, the authors proposed an \gls{UDA} method to align the feature extractor from a source dataset to a specific target dataset. This is done within the context of COVID-19 detection using chest X-ray images.  The feature extractor was originally trained with source data. Later, the feature extractor is  aligned by using both labelled and unlabelled data from the target dataset. The feature extractor alignment procedure basically consists in an adversarial training step using the aforementioned datasets. As a disadvantage of such method, the feature extractor needs to be trained with labelled source data (as usual in supervised learning). Hence a large number of labels is needed. Also,  the feature extractor alignment process can be considered  to be expensive, as an adversarial loss function needs to be optimized.

\section{Proposed method}\label{sec:Proposed_method}

\subsection{\gls{SSDL} with MixMatch}

In this work, we explore the usage of  MixMatch  as an \gls{SSDL} method, therefore, we describe it as follows. For more details please refer to \cite{berthelot2019mixmatch}. As previously mentioned, MixMatch combines both pseudo label and consistency regularization \gls{SSDL}. In such context, a pseudo-label $\widehat{\boldsymbol{y}}{}_{j}$ is estimated for each unlabelled observation $\boldsymbol{x_{j}}$ in $X_u$. It corresponds to the the mean model output of a transformed input $\boldsymbol{x'_{j}}$, using $K$ number of different transformations, such as flips and rotations \cite{berthelot2019mixmatch}. Each pseudo-label $\widehat{\boldsymbol{y}}$ is sharpened using a  temperature parameter $T$ \cite{berthelot2019mixmatch}. Also, a  simple data augmentation approach is implemented, by linearly combining unlabelled and labelled observations, through the usage of the MixUp algorithm \cite{zhang2017mixup}. 

The pseudo-labels are used in the MixMatch loss function, which combines a supervised and unsupervised loss terms. In this work, the well-known cross-entropy function is used as a supervised loss term. As for the unsupervised loss term, we used the previously implemented Euclidian distance loss in \cite{berthelot2019mixmatch}. The Euclidian distance measures the distance between the current model output and its pseudo-label, for the unlabelled observations. This loss term is weighed by the unsupervised learning coefficient $\gamma$. In this work, we used the MixMatch hyperparameters recommended in \cite{berthelot2019mixmatch}, of $K=2$, and $T=0.25$. As for the unsuperivsed coefficient, a value of $\gamma=200$ is used, given our empirical test results. 

\subsection{Harm coefficient estimation for unlabelled observations}
Interesting results were yielded in \cite{calderonramirez2020mixmood,calderon2021more}, where the authors found an strong correlation between the feature-density based distances and the MixMatch's accuracy. Based upon it, we propose to estimate how harmful an individual unlabelled observation might be towards the MixMatch's level of accuracy. We refer to this operator as the \gls{SSDL} harm coefficient $\mathcal{H}\left(\boldsymbol{x}^{u}_{j}\right)$, where $\boldsymbol{x}^{u}_{j}\in S_u$.
We aim to implement a simple and computationally inexpensive method to filter \gls{OOD} data in the unlabelled dataset, This is done in order to decrease the distribution mismatch between $S_u$ and $S_l$. 

As mentioned in Section \ref{sec:stateoftheart}, using different unlabelled data sources might increase the chance of violating the clustered-data/low-density separation assumption. This is particularly the case given the potential distribution mismatch between the labelled and unlabelled datasets. Therefore, our proposed method aims to discard harmful observations that might create wrong low density regions to build the manifold and/or sparser sample clusters for each category.
In a real-world scenario for \gls{OOD} filtering, \gls{DNN}s are fed with high resolution images, frequently with images from the same domain (chest X-ray images in our case). This contrasts with the usual settings of the methods discussed in Section \ref{sec:stateoftheart}. As previously discussed,   benchmarking in the literature have been usually performed with small resolution images and with relatively not very difficult \gls{OOD}  detection challenges (i.e distinguishing between CIFAR-10 and MNIST images). We aim to further test real-world distribution mismatch conditions in a medical imaging analysis application such as the COVID-19 detecion using chest X-ray images.

In this work, we propose to use the feature density of a labelled dataset $S_l$, to weigh how harmful could be to include an unlabelled observation $\boldsymbol{x}^u_j$ in the unlabelled dataset $S_u$. This is done witin the context of training a model using the \gls{SSDL} algorithm known as MixMatch. This harmful coefficient is represented as $\mathcal{H}\left(\boldsymbol{x}_{j}^{u}\right)$. We test two different variations to estimate $\mathcal{H}\left(\boldsymbol{x}_{j}^{u}\right)$. The first one consists in a non-parametric estimation of the feature density through an histogram calculation. The second variation assumes a Gaussian distribution of the feature space, by using a Mahalanobis distance. We use  a generic feature-space built from a pre-trained image-net model, to keep the computational cost  of the proposed method low. For all the tested configurations, we only use the features of the final convolutional layer. Computational resource restrictions for solving a real-world problem in medical imaging  makes very expensive to use all the features extracted in the different layers as done in \cite{lee2018simple}.
The procedure to calculate the harm coefficient using both methods, is depicted as follows:
\begin{enumerate}
    \item  For all of the input observations $\boldsymbol{x}^{l}_j\in S_l$, with $\boldsymbol{x}^{l}_j \in \mathbb{R}^{n}$, being $n$  the input space  dimensionality, using the feature extractor $f$, we calculate its feature vector as $ \boldsymbol{h}^{l}_{j}=f\left(\boldsymbol{x}^{l}_{j}\right)$. 
    \item The feature vector $\boldsymbol{h}^{l}_{j}  \in \mathbb{R}^{n'}$ has dimension $n'$ , with $n' < n$. For instance, a given feature extractor $f$ using the Imagenet pretrained  Wide-ResNet architecture, yields $n'=512$ features. For architectures such as densenet that might yield larger feature arrays in its final convolutional layer, we sub-sampled it to keep it in $n'=1024$ features, using an average pooling operation.   This yields a feature set $H_{l}$.
    \item For the \gls{FH} method, we perform the following steps:
    \begin{enumerate}
         \item  For each dimension $r=1,...,n'$ in the feature space, we compute its normalized histogram to approximate the density functions $\widetilde{p}_{r}^{l}$, in the sample $H_{l}$. This yields the set of approximated feature density functions:
            \begin{equation}
                \widetilde{P}^{l}=\left\{ \widetilde{p}_{1}^{l},\ldots,\widetilde{p}_{n'}^{l}\right\} 
            \end{equation}
        \item Using the approximated feature densities in $\widetilde{P}^{l}$, we estimate our \gls{SSDL} harm coefficient $\mathcal{H}\left(\boldsymbol{x}^{u}_{j}\right)$, for an unlabelled observation  in the following steps $\boldsymbol{x}^{u}_{j}$.
        \item Calculate the features for each unlabelled observation as $\boldsymbol{h}_{j}^{u}=f\left(\boldsymbol{x}_{j}^{u}\right)$, for each dimension in $\boldsymbol{h}_{j}^{u}\in\mathbb{R}^{n'}$, 
    
    \item The total likelihood calculation within the density function approximation set $\widetilde{P}^{l}$ assumes that each dimension is statistically independent. Thus: 
    \begin{equation}
    \prod_{r=1}^{n'}p_{r}^{l}\left(h_{j,r}^{u}\right).
    \end{equation}
    \item To avoid under-flow, we calculate the negative logarithm of the likelihood, and use it as the harm coefficient: \begin{equation}
     \mathcal{H}\left(\boldsymbol{x}_{j}^{u}\right)=-\sum_{r=1}^{n'}\ln\left(p_{r}^{l}\left(h_{j,r}^{u}\right)\right).
    \end{equation}

    \end{enumerate} 
   \item For the Mahalanobis based filtering, we perform the following steps:
   \begin{enumerate}
       \item Calculate the covariance matrix $\Sigma$ from the features set $H_{l}$, and the sample mean from the features set $\overline{h}_{l}$.
       \item Calculate the features for each unlabelled observation as $\boldsymbol{h}_{j}^{u}=f\left(\boldsymbol{x}_{j}^{u}\right)$.
       \item Compute the harm coefficient as:
       \begin{equation}
           \mathcal{H}\left(\boldsymbol{x}_{j}^{u}\right)=\left(\overline{\boldsymbol{h}}_{l}-\boldsymbol{h}_{j}^{u}\right)^{T}\Sigma^{-1}\left(\overline{\boldsymbol{h}}_{l}-\boldsymbol{h}_{j}^{u}\right).
       \end{equation}
   \end{enumerate}
\end{enumerate}

The harm coefficient $  \mathcal{H}\left(\boldsymbol{x}_{j}^{u}\right)$ can be used to discard the observations with high values, or to weigh them in case an online semi-supervised per-observation weighting is implemented.  In this work, we test the impact of the distribution mismatch between the labelled target and unlabelled source datasets, $D^{l}_t$ and $D^{u}_s$, respectively, in the accuracy of the \gls{SSDL} MixMatch algorithm. Later, we test the impact of the proposed feature based \textit{harm coefficient} to eliminate potentially harming observations from the unlabelled dataset. This was done to assess the accuracy of the model using the filtered unlabelled  dataset $D^{u}_s$. This way, we can assess in a controlled setting the impact of the distribution rectification procedure, implemented through a data filtering process.

\section{Datasets}\label{sec:datasets}

In this work, we explore the sensitivity to distribution mismatch between $S_u$ and $S_l$ of a \gls{SSDL} COVID-19 detection system using chest X-ray images. Therefore, we use different data sources for chest X-ray images for both COVID-19$^{+}$ (positive COVID-19) and    COVID-19$^{-}$ (no pathology chest X-ray observations). For  COVID-19$^{+}$ cases we use the open dataset made available by Dr. Cohen in \cite{cohen2020covid}. This dataset  is composed of  105 COVID-19$^{+}$ images at the time of writing this work. The observations were sampled from different journal websites like the Italian Society of Medical and Interventional Radiology and  \url{radiopaedia.org}, and more recent publications in the field. In this work we used  COVID-19$^{+}$ observations, discarding images related to \gls{MERS}, \gls{ARDS} and \gls{SARS}.

\begin{figure}
\caption{  Row 1, column 1: a COVID-19$^{+}$ observation from \cite{cohen2020covid}, row 1, column 2:  a  COVID-19$^{-}$ observation from the Chinese dataset \cite{kermany2018identifying}, row 2, column 1: ChestX-ray8  COVID-19$^{-}$ image \cite{irvin2019chexpert}, row 2, column 2: Indiana dataset COVID-19$^{-}$  sample image \cite{demner2016preparing}. The bottom image corresponds to a sample image from the Costa Rica dataset \cite{calderon2020correcting}.  As it can be seen, images from the Costa Rica dataset include a black frame.
\\
\label{fig:sample_images}}
\centering{}
\includegraphics[scale=0.24]{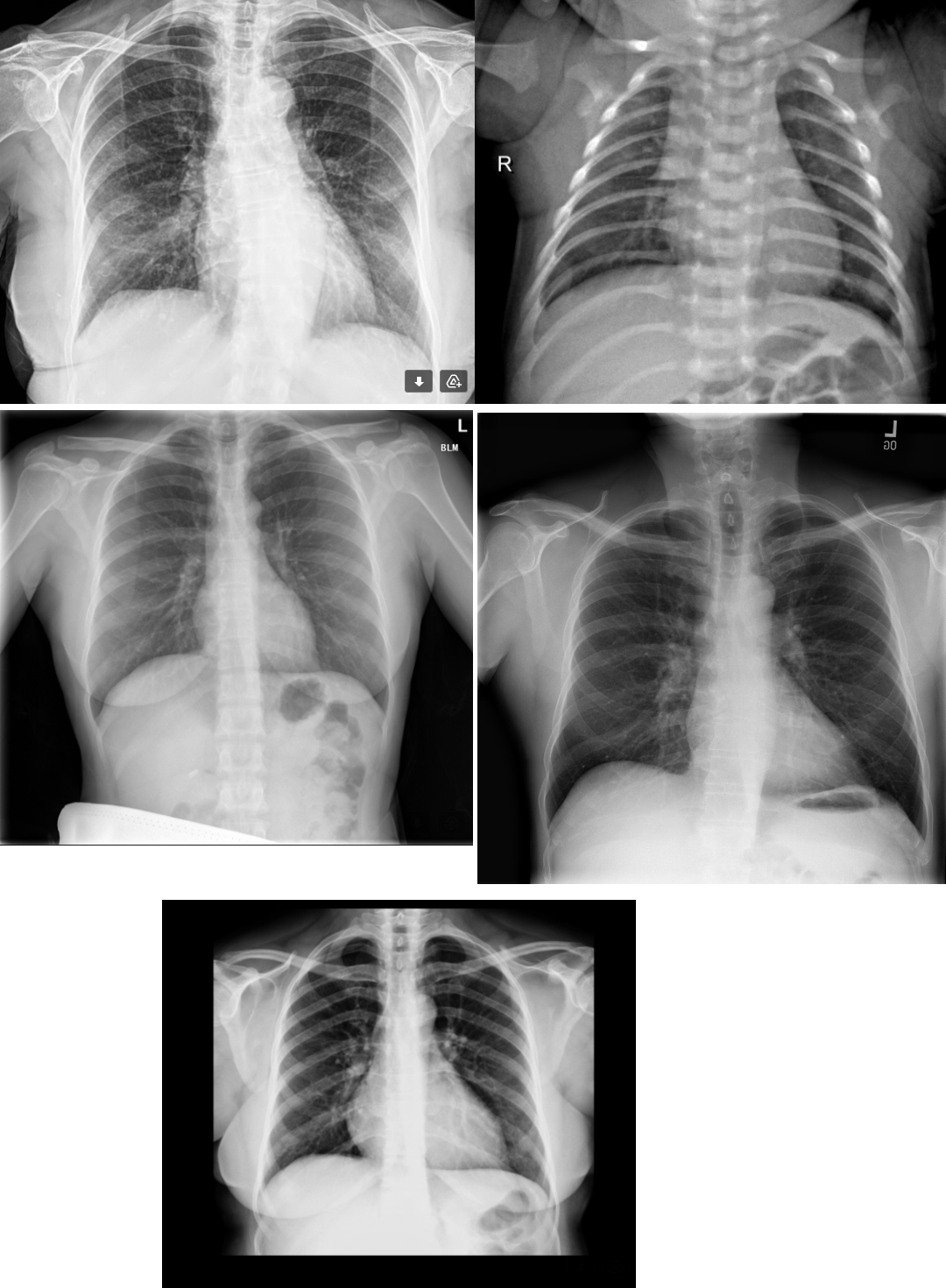}

\end{figure}

The images present varying resolutions from $400 \times 400$ up to $2500 \times 2500$ pixels. As for COVID-19$^{-}$ observations, we used four different data-sources.  Table \ref{tab:datset_description} summarizes the  COVID-19$^{-}$ cases data sources.   Figure \ref{fig:sample_images} shows observations for each one of the data sources used in this work.  The datasets were randomly augmented with flips and rotations. No random crops were used to avoid discarding important regions in the images. 

In this first set of experiments, we evaluate the impact of \gls{OOD} on data with different unlabelled data sources and different degrees of \textit{contamination}.  We simulate the following scenario: A small labelled target dataset $D^{t}_l$ (with $n_l=20$ and $n_l=40$ observations) is provided with a partition of the  observations of the COVID-19$^+$ taken from Dr. Cohen's dataset and the COVID-19$^-$ cases of the Indiana Chest X-ray  dataset, described in Table \ref{tab:datset_description}. A larger number of 142 unlabelled observations is also available, to be used in the harm coefficient estimations methods.  This can be thought as the target labelled dataset with limited labels which is accessible in a real-world application from the clinic/hospital where the model is intended to be deployed. 

For the unlabelled dataset, different partitions of COVID-19$^-$ cases the chest X-ray  data sources  described in Table \ref{tab:datset_description}. This simulates the usage of different sources of unlabelled datasets $D^{s}_u$, taken from different hospitals/clinics. All the unlabelled observations are COVID-19$^-$, to enforce a prior probability shift (label imbalance). As in our  preliminar tests, the worst performing unlabelled dataset $D^{s}_u$ dataset is the Costa Rican dataset described in Table \ref{tab:datset_description}, we used it to create different combinations with the rest of datasets. All of these are depicted in Table \ref{tab:accuracy_baseline_alexnet}.      A total of $n_u = 90$ unlabelled observations were picked from such datasests with different combinations. Using different data sources for the unlabelled dataset, can help to assess the impact of a distribution mismatch between $S_u$ and $S_l$.  
    
As for the test dataset, it consists in  another partition of the target dataset which includes the  COVID-19$^+$ dataset, along with another partition of the Indiana Chest X-ray  dataset (COVID-19$^-$). Both are the same size. This yiels a completely balanced test setting.  We used a total of  $n_t = 62$ observations, drawn from the same target dataset (31 observations per class). The test data comes from the distribution of the labelled data with no contamination. This simulates the case where the labelled data comes from the target dataset distribution. Both unlabelled and labelled datasets were standardised, given that the authors in \cite{calderon2020mixmood} found that normalisation is important in semi-supervised learning. 

\begin{table*}[]
\caption{COVID-19$^{-}$ observation sources description used in this work.}\label{tab:datset_description}
\center
\begin{tabular}{c|c|c|c|c}
\textbf{}                   & \textbf{Costa Rican dataset}  & \textbf{Chinese dataset} & \textbf{ChestX-ray8 dataset}  & \textbf{Indiana dataset}  \\ \hline
No. of patients             & 105                          &   5856                       & 65240                        & 4000                         \\
Patient's age range (years) & 7-86                         &   children                       & 0-94                         &     adults                     \\
No. of  obs.                &   105                           & 5236                     & 224316                       & 8121                     \\
Hospital/clinic             & Clinica Chavarria            &  No info.                        & Stanford Hospital            & Indiana Network          \\
                            &                              &                          &                              & for Patient Care         \\
Im. resolution              & $1907\times1791$             & $1300\times600$          &     $1024\times1024$                         & $1400\times1400$         \\
Reference                   &     \cite{calderon2020correcting}                         &     \cite{kermany2018identifying}                     &  \cite{irvin2019chexpert}                            &         \cite{demner2016preparing}                 \\
                            &                              &                          &                              &                         
\end{tabular}
\end{table*}

\section{Experiments}

 \gls{TB-1}    is designed to assess the effect of on MixMatch's accuracy of using different unlabelled datasets  $D^{s}_u$ with a target labelled dataset  $D^{t}_l$. This test-bed recreates different distribution mismatch conditions between $D^{s}_u$ and  $D^{t}_l$.  The Costa Rican dataset acts as a source of \gls{OOD} data, as it yielded the lowest accuracy when used as $D^{s}_u$ for MixMatch, among the empirically tested unlabelled datasources. We combine the aforementioned data sources with the Costa Rican dataset. This helps enforce different distribution mismatch settings. 
 
 In the \gls{TB-1.1}, the first  sub-experiment defined within the \gls{TB-1}, we measure MixMatch's accuracy using a densenet model, with feature extractor fine-tuning and without it. We aim to measure if there is a significant accuracy gain of fine-tuning the feature extractor during training. Table \ref{tab:accuracy_baseline_densenet_no_finetuning} shows the results of performing MixMatch's training  without feature extractor fine-tuning, while Table \ref{tab:accuracy_baseline_finetuned_densenet} shows the results with it.

Additionally, we devised a \gls{TB-1.2}, where the baseline results obtained in this MixMatch accuracy baseline test-bed in Tables \ref{tab:accuracy_baseline_densenet_no_finetuning} and \ref{tab:accuracy_baseline_alexnet} are  correlated with the cosine \gls{DeDiM}s between each $D^{s}_u$ and  $D^{s}_u$. This is  measured as proposed in \cite{calderon2020mixmood}, and represented as $d_C(D^{s}_u, D^{t}_l)$. For this experiment, we tested an alexnet's model feature extractor, given its low computational cost.  We implemented the cosine dataset \gls{DeDiM} with a batch dataset size of $n_b =40$, with 10 batches of random samples. The same batches were used to test the different configurations. Similar to the proposed harm coefficient estimation methods, we used a generic Imagenet pre-trained feature extractor to build the feature density estimations, as proposed in \cite{calderon2020mixmood}. The \gls{DeDiM} results are linearly  correlated using a Pearson coefficient in Table \ref{tab:Pearson}.

Finally, \gls{TB-2} aims to assess MixMatch's accuracy results when implementing the proposed methods in this work to filter the \gls{OOD} observations, against two popular output based \gls{OOD} filtering methods: the \gls{MCD} and Softmax based \gls{OOD} filters. In this test bed, we measure MixMatch's accuracy through the four different filtered datasets, testing both alexnet and densenet as a model. We also tested the model with  $n_l=20$ and $n_l=40$ labels.   The results using the proposed feature histograms and Mahalanobis distance for each generated unlabelled data source $D^{s}_u$ are depicted in Tables \ref{tab:alexnet_filtered_fh_mahalanobis_results} and \ref{tab:densenet_filtered_fh_mahalanobis_results}, for the alexnet and the densenet models, respecitvely. To filter possible \gls{OOD} observations, we eliminated the same percent of contaminated observations using the Costa Rican dataset (i.e, if the Chinese dataset was contaminated with 35\% of observations with the Costa Rican dataset, we eliminated 35\% of the observations with the highest harm coefficient, and so on). We leave the problem of defining the right harm coefficient threshold out of this study. 

In all test beds, the MixMatch algorithm is tested with a densenet and alexnet models, using the recommended parameters in \cite{berthelot2019mixmatch}, along with an unsupervised regularization term coefficient of 200. As for model training, we use the one-cycle policy implemented in the FastAI library, with a weight decay of 0.001,    This way we can measure MixMatch's behaviour with models with different depth and architecture.  For each configuration, we trained the model with 10 runs, using a different random data partition for training and test, for 50 epochs.

\begin{table}[]
\caption{ \gls{TB-1.1} results: Accuracy of a Densenet model trained with MixMatch with different $D^{s}_u$ datasets. The unlabelled datasets Chest-Xray8, Costa Rican and Chinese datasets include only COVID-19$^{-}$ observations. No use of a fine-tuned feature extractor. } \label{tab:accuracy_baseline_densenet_no_finetuning}
\begin{tabular}{l|c|c}
Dataset                      & \multicolumn{1}{l|}{$n_l=40$} & \multicolumn{1}{l}{ $n_l=20 $} \\ \hline
Supervised            & $0.851\pm0.037$                               & $0.803\pm0.039$                                 \\
Indiana (with COVID-19$^{+}$ \cite{cohen2020covid})            &            $0.891\pm0.047$                       & $0.875\pm0.04$                                 \\
 China            &  $0.735\pm0.0621$                                & $0.722\pm0.054$                                 \\
Costa Rica       & $0.493\pm0.014$                                 &      $0.511\pm0.029$                       \\
ChestX-ray8              & $0.825\pm0.061$                                  &  $0.795\pm0.052$                                \\
ChestX-ray8 65\% - Costa Rica 35\%   & $0.579\pm0.115$                           &$0.582\pm0.067$                                 \\
ChestX-ray8 35\% - Costa Rica 65\%   & $0.5\pm0.001$                            & $0.503\pm0.009$                                 \\

China 65\% - Costa Rica 35\% &$0.588\pm0.066$                                   &                                   $0.559\pm0.067$ \\
China 35\% - Costa Rica 65\%                                  &$0.498\pm0.004$      &  $0.508\pm0.024$     \\

Indiana 65\% - Costa Rica 35\% &$0.504\pm0.014$                                   & $0.553\pm0.062$                                 \\
Indiana 35\% - Costa Rica 65\%                                  & $0.501\pm0.004$    & $0.5\pm0.001$      
\end{tabular}
\end{table}

\begin{table}[]
\caption{\gls{TB-1.1} results: Accuracy of a Densenet model trained with MixMatch with different $D^{s}_u$ datasets. The unlabelled datasets Chest-Xray8, Costa Rican and Chinese datasets include only COVID-19$^{-}$ observations.  Using the fine-tuned feature extractor.} \label{tab:accuracy_baseline_finetuned_densenet}
\begin{tabular}{l|c|c}
Dataset                      & \multicolumn{1}{l|}{$n_l=40$} & \multicolumn{1}{l}{ $n_l=20 $} \\ \hline
Supervised            & $0.852\pm0.045$                               & $0.795\pm0.005$                                  \\
Indiana (with COVID-19$^{+}$ \cite{cohen2020covid})            &            $0.892\pm0.044$                        & $0.885\pm0.039$                                  \\
 China            &  $0.733\pm0.043$                                 & $0.709\pm0.059$                                  \\
Costa Rica       & $0.498\pm0.004$                                 &      $0.501\pm0.016$                        \\
ChestX-ray8              & $0.804\pm0.061$                                   &  $0.793\pm0.044$                                 \\
ChestX-ray8 65\% - Costa Rica 35\%   & $0.598\pm0.1$                            &$0.591\pm0.105$                                  \\
ChestX-ray8 35\% - Costa Rica 65\%   & $0.501\pm0.004$                             & $0.488\pm0.033$                                  \\

China 65\% - Costa Rica 35\% &$0.593\pm0.057$                                    &                                   $0.614\pm0.0926$  \\
China 35\% - Costa Rica 65\%                                  &$0.514\pm0.055$       &  $0.496\pm0.022$      \\

Indiana 65\% - Costa Rica 35\% &$0.516\pm0.048$                                    & $0.535\pm0.047$                                  \\
Indiana 35\% - Costa Rica 65\%                                  &  $0.508\pm0.016$    & $0.501\pm0.011$       
\end{tabular}
\end{table}

\begin{table}[]
\caption{\gls{TB-1.1} results: Accuracy of a Alexnet model trained with MixMatch with different $D^{s}_u$ datasets. The unlabelled datasets Chest-Xray8, Costa Rican and Chinese datasets include only COVID-19$^{-}$ observations.   } \label{tab:accuracy_baseline_alexnet}
\begin{tabular}{l|c|c}
Dataset                      & \multicolumn{1}{l|}{$n_l=40$} & \multicolumn{1}{l}{ $n_l=20 $} \\ \hline
Supervised            & $0.785\pm0.038$                               & $0.809\pm0.085$                                  \\
Indiana (with COVID-19$^{+}$ \cite{cohen2020covid})            &            $0.782\pm0.039$                        & $0.75\pm0.06$                                  \\
 China            &  $0.648\pm0.0247$                                 & $0.659\pm0.033$                                  \\
Costa Rica       & $0.501\pm0.001$                                 &      $0.5\pm0.001$                        \\
ChestX-ray8              & $0.72\pm0.076$                                   &  $0.71\pm0.074$                                \\
ChestX-ray8 65\% - Costa Rica 35\%   & $0.711\pm0.083$                          &$0.66\pm0.11$                                    \\
ChestX-ray8 35\% - Costa Rica 65\%   &   $0.516\pm0.022$                             &     $0.511\pm0.016$                            \\

China 65\% - Costa Rica 35\% &$0.701\pm0.055$                                    &                              $0.688\pm0.084$      \\
China 35\% - Costa Rica 65\%                                  & $0.53\pm0.023$      &     $0.528\pm0.019$    \\

Indiana 65\% - Costa Rica 35\% &$0.532\pm0.024$                                    & $0.559\pm0.059$                                  \\
Indiana 35\% - Costa Rica 65\%                                  & $0.501\pm0.001$     & $0.503\pm0.009$       
\end{tabular}
\end{table}

\begin{table}[]
\caption{\gls{TB-1.2} results: Cosine \gls{DeDiM} distance, using 10 different batches of 80 observations, between the labelled and unlabelled datasets, $S_l$ and $S_u$, respectively. Using Alexnet, to keep computing cost low.  }\label{tab:Distances}
\center
\begin{tabular}{l|l}
\textbf{Dataset}                   & $d(S_l,S_u)$   \\ \hline
China                              & $2.06\pm0.11$ \\
Costa Rica                         & $30.9\pm0.4$ \\
ChestX-ray8                        & $1.04\pm0.27$ \\
ChestX-ray8 65\% - Costa Rica 35\% & $3.95\pm0.94$ \\
ChestX-ray8 35\% - Costa Rica 65\% & $11.84\pm0.94$  \\
China 65\% - Costa Rica 35\%       & $5.74\pm0.79$  \\
China 35\% - Costa Rica 65\%       & $14.85\pm0.0$ \\
Indiana 65\% - Costa Rica 35\%    & $6.33\pm0.3$ \\
Indiana 35\% - Costa Rica 65\%    & $16.61\pm0.3$
\end{tabular}
\end{table}

\begin{table}[]
\caption{\gls{TB-1.2} test results:  Pearson coefficient between the accuracy and the calculated divergences.}\label{tab:Pearson}
\center
\begin{tabular}{l|l|l}
\textbf{SSDL model} & $n_l$ & \textbf{Pearson  coefficient} \\ \hline
Alexnet             & 20    & -0.798                        \\
                    & 40    & -0.75                         \\ \hline
Densenet            & 20    & -0.665                        \\
                    & 40    & -0.662                       
\end{tabular}
\end{table}

\begin{table*}[]
\center
\caption{Accuracy of a Alexnet model trained with MixMatch, with the filtered datasets using the harm coefficient with the two output-based methods: \gls{MCD} and Softmax. The percentage of discarded observations is the same of the amount of Costa Rican observations. }\label{tab:alexnet_filtered_softmax_mcd}
\begin{tabular}{l|c|c|c|c}
\multicolumn{1}{c|}{\textbf{}}     & \multicolumn{2}{c|}{$n_l=40$} & \multicolumn{2}{c}{$n_l=20$} \\
\multicolumn{1}{c|}{Dataset}       & Acc.  Softmax    & Acc.  MCD  & Acc.  Softmax    & Acc.  MCD  \\ \hline
ChestX-ray8 35\% - Costa Rica 65\% & $0.532\pm0.059$    & $0.506\pm0.012$  & $0.52\pm0.038$  & $0.5\pm0.002$  \\
ChestX-ray8 65\% - Costa Rica 35\% & $0.582\pm0.096$   & $0.567\pm0.067$  & $0.579\pm0.096$  & $0.558\pm0.067$  \\
China 35\% - Costa Rica 65\%       & $0.514\pm0.04$  & $0.503\pm0.009$  & $0.525\pm0.077$    & $0.509\pm0.02$  \\
China 65\% - Costa Rica 35\%       & $0.591\pm0.096$  & $0.579\pm0.076$  & $0.585\pm0.096$    & $0.567\pm0.051$  \\
Indiana 35\% - Costa Rica 65\%     & $0.503\pm0.009$        & $0.503\pm0.006$  & $0.506\pm0.019$        & $0.509\pm0.014$  \\
Indiana 65\% - Costa Rica 35\%     & $0.574\pm0.078$        & $0.544\pm0.032$  &$0.551\pm0.054$         & $0.543\pm0.042$ 
\end{tabular}
\end{table*}

\begin{table*}[]
\center
\caption{Accuracy of a Alexnet model trained with MixMatch, with the filtered datasets using the harm coefficient with the two proposed feature density based methods: \gls{FH} and the Mahalanobis based filter. The percentage of discarded observations is the same of the amount of Costa Rican observations.}\label{tab:alexnet_filtered_fh_mahalanobis_results}
\begin{tabular}{l|c|c|c|c}
\multicolumn{1}{c|}{}        & \multicolumn{2}{c|}{$n_l=40$}                                                    & \multicolumn{2}{c}{$n_l=20$}                                                    \\
\multicolumn{1}{c|}{Dataset} & Acc.  FD & Acc.  Mahalanobis & Acc.  FD & Acc.  Mahalanobis \\ \hline
ChestX-ray8 35\% - Costa Rica 65\%   & $0.709\pm0.084$                    & $0.727\pm0.078$                                   &   $0.682\pm0.09$                 & $0.685\pm0.089$                                     \\
ChestX-ray8 65\% - Costa Rica 35\%   & $0.732\pm0.064$                    &  $0.7612\pm0.049$                                    &        $0.717\pm0.08$            & $0.709\pm0.09$                                      \\
China 35\% - Costa Rica 65\% & $0.683\pm0.065$                    & $0.708\pm0.07$                                    & $0.667\pm0.078$                   & $0.667\pm0.09$                                     \\
China 65\% - Costa Rica 35\% & $0.693\pm0.044$                     & $0.695\pm0.079$                                     &          $0.687\pm0.078$         & $0.674\pm0.072$                 
\\

Indiana 35\% - Costa Rica 65\% & $0.732\pm0.052$                    & $0.711\pm0.032$                                    & $0.703\pm0.1$                   & $0.719\pm0.09$                                     \\
Indiana 65\% - Costa Rica 35\% & $0.719\pm0.058$                     & $0.748\pm0.059$                                    &           $0.709\pm0.093$          & $0.711\pm0.09$   
\end{tabular}
\end{table*}

\begin{table*}[]
\center
\caption{Accuracy of a Densenet model trained with MixMatch, with the filtered datasets using the harm coefficient with the two output-based methods: \gls{MCD} and Softmax. The percentage of discarded observations is the same of the amount of Costa Rican observations. }\label{tab:densenet_filtered_Softmax_MCD_results}
\begin{tabular}{l|c|c|c|c}
\multicolumn{1}{c|}{\textbf{}}     & \multicolumn{2}{c|}{$n_l=40$} & \multicolumn{2}{c}{$n_l=20$} \\
\multicolumn{1}{c|}{Dataset}       & Acc.  Softmax    & Acc.  MCD  & Acc.  Softmax    & Acc.  MCD  \\ \hline
ChestX-ray8 35\% - Costa Rica 65\% & $0.5\pm0.001$    & $0.5\pm0.001$  & $0.488\pm0.025$  & $0.529\pm0.077$  \\
ChestX-ray8 65\% - Costa Rica 35\% & $0.543\pm0.09$   & $0.537\pm0.11$  & $0.543\pm0.095$  & $0.498\pm0.004$  \\
China 35\% - Costa Rica 65\%       & $0.498\pm0.004$  & $0.5\pm0.001$  & $0.49\pm0.04$    & $0.496\pm0.009$  \\
China 65\% - Costa Rica 35\%       & $0.517\pm0.029$  & $0.501\pm0.004$  & $0.5\pm0.007$    & $0.504\pm0.01$  \\
Indiana 35\% - Costa Rica 65\%     & $0.499\pm0.001$        & $0.5\pm0.001$  & $0.48\pm0.036$        & $0.496\pm0.009$  \\
Indiana 65\% - Costa Rica 35\%     & $0.5\pm0.001$        & $0.501\pm0.008$  & $0.497\pm0.$        & $0.503\pm0.0173$ 
\end{tabular}
\end{table*}

\begin{table*}[]
\center
\caption{Accuracy of a Densenet model trained with MixMatch, with the filtered datasets using the harm coefficient with the two proposed feature density based methods: \gls{FH} and the Mahalanobis based filter. The percentage of discarded observations is the same of the amount of Costa Rican observations.}\label{tab:densenet_filtered_fh_mahalanobis_results}
\begin{tabular}{l|c|c|c|c}
\multicolumn{1}{c|}{}        & \multicolumn{2}{c|}{$n_l=40$}                                                    & \multicolumn{2}{c}{$n_l=20$}                                                    \\
\multicolumn{1}{c|}{Dataset} & Acc.  FD & Acc.  Mahalanobis & Acc.  FD & Acc.  Mahalanobis \\ \hline
ChestX-ray8 35\% - Costa Rica 65\%   & $0.691\pm0.10$                    & $0.769\pm0.048$                                     &    $0.683\pm0.105$                 & $0.779\pm0.025$                                     \\
ChestX-ray8 65\% - Costa Rica 35\%   & $0.717\pm0.091$                    &  $0.811\pm0.049$                                    &         $0.695\pm0.1$            & $0.783\pm0.049$                                      \\
China 35\% - Costa Rica 65\% & $0.794\pm0.036$                    & $0.795\pm0.053$                                     & $0.787\pm0.048$                    & $0.769\pm0.076$                                     \\
China 65\% - Costa Rica 35\% & $0.788\pm0.056$                     & $0.812\pm0.05$                                     &           $0.774\pm0.053$          & $0.798\pm0.036$                 
\\

Indiana 35\% - Costa Rica 65\% & $0.758\pm0.047$                    & $0.729\pm0.035$                                     & $0.727\pm0.0512$                    & $0.714\pm0.046$                                     \\
Indiana 65\% - Costa Rica 35\% & $0.737\pm0.049$                     & $0.762\pm0.055$                                     &           $0.703\pm0.055$          & $0.722\pm0.032$  
\end{tabular}
\end{table*}

\section{Results Analysis}

As for the results in \gls{TB-1.1}, depicted in Table \ref{tab:accuracy_baseline_densenet_no_finetuning}, we can see a very strong influence of the unlabelled data source $D^{s}_u$ in the accuracy of the \gls{SSDL} MixMatch algorithm.  Training the model with the Indiana dataset including also COVID-19$^{+}$ observations, yields the highest accuracy, with around 0.89, higher than the supervised model. From there, using the ChestX-ray8 as $D^{s}_u$, yields an accuracy of 0.825, followed by the usage of the Chinese dataset as $D^{s}_u$, accuracy wise. Using the Costa Rican dataset as $D^{s}_u$ yields the lowest accuracy, with close to 0.493.  \textit{Contaminating} the ChestXray8, Chinese and Indiana dataset (with only COVID-19$^{-}$ observations for all of them), yields a lower accuracy with an increasing degree of contamination. 
As for the impact of fine-tuning the feature extractor, there is no statistical significant difference of performing it, when comparing the results in Tables \ref{tab:accuracy_baseline_densenet_no_finetuning} and \ref{tab:accuracy_baseline_finetuned_densenet}. This suggests that using an image-net pre-trained feature extractor for harm coefficient estimation is justifiable.

  Regarding \gls{TB-2} results, when comparing the accuracy yielded by MixMatch for each tested  $D^{s}_u$ with the calculated inter-dataset cosine \gls{DeDiM}s in Table \ref{tab:Distances}, we can see an interesting relationship. The Costa Rican dataset and heavily contaminated $D^{s}_u$ data sources present the highest distances. For instance, the Chinese dataset contaminated with a degree of 65\% with the Costa Rican dataset, presents a distance of 50.93 with the labelled dataset $D^{s}_u$, similar to the inter-dataset distance to the Costa Rican dataset of 57.19 (the $D^{s}_u$ with the highest distance to $D^{t}_l$). We can see how using both of the aforementioned $D^{s}_u$ datasets, yield very low MixMatch accuracy. This behaviour is summarized in the obtained Pearson coefficients depicted in Table \ref{tab:Pearson}, with a very high lineal correlation, of around 78\%   for the tested variations. The correlation is still high for the semi-supervised densenet model behaviour with the dataset distances, using a generic imagenet pre-trained alexnet model.    This suggests that the usage of the feature density can bring useful information to preserve or discard an unlabelled observation in a $D^{s}_u$. 

Regarding the results of \gls{TB-2}, Tables \ref{tab:densenet_filtered_fh_mahalanobis_results} and \ref{tab:alexnet_filtered_fh_mahalanobis_results} show the accuracy of MixMatch yielded when filtering the unlabelled datasets with the proposed \gls{FH} and Mahalanobis methods, for both tested models (alexnet and densenet, respectively). For both proposed methods, we can see how filtering potentially harming observations from the unlabelled dataset  increases MixMatch's accuracy significantly, when compared to the baseline accuracies in Tables \ref{tab:accuracy_baseline_alexnet} and \ref{tab:accuracy_baseline_densenet_no_finetuning}, for both tested models.  For instance, when using the densenet model with $n_l = 40$, the ChestX-ray8 dataset contaminated with 35\% and 65\% with the \textit{Costa Rica} dataset, increases its accuracy from 0.579 to 0.78 and 0.5 to 0.79, respectively, when filtering harmful observations with the Mahalanobis method. This can be  seen in both Tables \ref{tab:accuracy_baseline_densenet_no_finetuning} and \ref{tab:densenet_filtered_fh_mahalanobis_results}. The usage of the \gls{FH} method yields also an important accuracy gain. In this case however, it is lower than the gains obtained with the Mahalanobis method.  The accuracy of the model trained with  $D^{s}_u$ using the ChestX-ray8 dataset with no contamination is almost restored, as MixMatch originally yielded 0.825. We have to consider that the filtered dataset is always smaller than the original unlabelled dataset. Despite this, the accuracy ends very close. Similarly, for the alexnet model with $n_l = 40$, the accuracy of using an \textit{Indiana} unlabelled dataset contaminated with 65\% of the \textit{Costa Rica} dataset is close to 50\%, according to Table \ref{tab:accuracy_baseline_alexnet}. However, after filtering out harmful unlabelled observations ends close to the 71\%, using both the \gls{FH} or the Mahalanobis method. 

When comparing the accuracy gain of using the feature histograms against the Mahalanobis distance based method, we can see a similar behaviour across almost all the tested unlabelled datasets $D^{s}_u$.  However, for the ChestX-ray8 dataset where the Mahalanobis based method yields statistically significant accuracy gains the the \gls{FH} approach, for the densenet model, as seen in Table \ref{tab:densenet_filtered_fh_mahalanobis_results}. This suggests that the feature distribution of the labelled dataset $D^{t}_l$ fits well with a Gaussian distribution, given the very slightly better results of the Mahalanobis method. The Mahalanobis based method is faster, as it only needs to compute a covariance matrix, when compared to the histogram based approach, which needs to build a feature histogram. This proved to be significantly slower in our tests.

As for the tested \gls{MCD} and Softmax baseline methods, popular in \gls{OOD} detection and uncertainty estimation, the results depicted in Tables \ref{tab:alexnet_filtered_softmax_mcd} and \ref{tab:densenet_filtered_Softmax_MCD_results}, for the alexnet and densenet models, show a very poor performance. The accuracy gains are negligible and sometimes the accuracy is diminished, when compared to the baseline results shown in Tables \ref{tab:accuracy_baseline_alexnet} and  \ref{tab:accuracy_baseline_densenet_no_finetuning}. Therefore, the usage of the feature density based methods for filtering potentially harmful unlabelled observations prove to be a significantly better approach.  Accuracy gains of up to 25\% with statistical significance in all the tested settings were obtained (using a Wilcoxon test with $p<0.1$), when using the feature density approaches over the tested output based ones. This can be seen when comparing the results for the proposed feature density techniques in Tables \ref{tab:alexnet_filtered_fh_mahalanobis_results} and \ref{tab:densenet_filtered_fh_mahalanobis_results}, with Tables \ref{tab:alexnet_filtered_softmax_mcd} and \ref{tab:densenet_filtered_Softmax_MCD_results}, for the both tested architectures alexnet and densenet, respectively.

\section{Conclusions}

In this work, we have analyzed the impact of the distribution mismatch between the labelled and the unlabelled dataset for training a \gls{SSDL} model, using the MixMatch algorithm. The setting assessed used medical imaging data, for COVID-19 detection. Assessing the impact of distribution mismatch between the unlabelled and labelled dataset for medical imaging applications is still an under-reported problem in the literature. 

In the first test-bed, we have assessed the impact of using different unlabelled data sources $D^{s}_u$, and quantitatively analyzed the distribution mismatch between them using \gls{DeDiM}s as a metric. The high linear correlation between the measured \gls{DeDiM}s and the MixMatch accuracy, suggests a strong influence of the  feature distribution mismatch between $D^{s}_u$ and $D^{t}_l$. In contexts where a decision must be made about what unlabelled data source $D^{s}_u$ must be used, from a set of possible unlabelled datasets, the \gls{DeDiM}s might be used as a quantitative prior method. Implementing the tested \gls{DeDiM}s requires no model training, as a generic pre-trained ImageNet model seems to be good enough to estimate the benefit of using a specific unlabelled dataset $D^{s}_u$, according to our results. Data quality metrics for deep learning models as argued in \cite{mendez2019using,balki2019sample} is an interesting path to develop further, as it might help to narrow the gap between research and real-world implementation of deep learning systems. For instance, building high quality datasets for training a semi-supervised model, or  assess the safety of using a deep learning model before hand, can benefit from quantitative data quality measures. We argue for the community to include robust data quality metrics in the deployment of  deep learning solutions.

To increase the robustness of the \gls{SSDL} model to the distribution mismatch, we tested different approaches to discard potentially harming unlabelled observations from the unlabelled dataset $D^{s}_u$. The tested setting can be considered to be closer to real-world settings, as images within the same domain were used as \gls{OOD} data contamination sources. This contrasts to the frequent \gls{OOD} detection benchmarks where images from very different dataset were used as \gls{OOD} data sources \cite{zisselman2020deep}.  Our approach is  data-oriented, as it modifies the original dataset in an explicit way by removing potentially harming unlabelled observations. We tested output based \gls{OOD} filtering techniques against our proposed feature density based approaches. Our proposed methods based on the feature densities built upon a pre-trained model with Imagenet, showed a large and significantly advantage over previous output based \gls{OOD} filtering methods. In the context of \gls{SSDL}, some approaches have relied in weighing each unlabelled observation using the output of the model, as in \cite{nair2019realmix}. According to our results, we argue that using the model's output might yield over-confident results to filter or weigh unlabelled observations. This is  widely known in \gls{OOD} detection literature \cite{le2018uncertainty}. Even ensemble based approaches like the tested \gls{MCD} method are not able to filter harming unlabelled observations, according to our test results. However, both feature density based approaches demonstrated a good performance on detecting harming unlabelled observations, almost recovering the original accuracy of the no contaminated datasets. The proposed methods can be deployed to correct and create more effective unlabelled datasets. Moreover both  proposed methods do not require any deep learning model training, making it cheap and reducing the carbon footprint of its implementation \cite{strubell2020energy}.

\bibliographystyle{plain}
\bibliography{bibliography}

\end{document}